\documentclass{article}

\usepackage{arxiv}

\usepackage[utf8]{inputenc} % allow utf-8 input
\usepackage[T1]{fontenc}    % use 8-bit T1 fonts
\usepackage{hyperref}       % hyperlinks
\usepackage{url}            % simple URL typesetting
\usepackage{booktabs}       % professional-quality tables
\usepackage{amsfonts}       % blackboard math symbols
\usepackage{nicefrac}       % compact symbols for 1/2, etc.
\usepackage{microtype}      % microtypography
\usepackage{multirow}
\usepackage{graphicx}
\usepackage{subcaption}

\usepackage{lipsum}

\usepackage{amsmath,amsthm,amssymb,amsfonts}
\usepackage{color}
\graphicspath{{figs}}

\title{Localization with Limited Annotation for Chest X-rays}

\author{
  Eyal Rozenberg \\
  Department of Computer Science \\
  Technion - Israel Institute of Technology \\
  Haifa, Israel \\
  \texttt{eyalr@campus.technion.ac.il} \\
  \And
  Daniel Freedman \\
  Google Research \\
  Haifa, Israel \\
  \texttt{danielfreedman@google.com} \\
  \And
  Alex Bronstein \\
  Department of Computer Science \\
  Technion - Israel Institute of Technology \\
  Haifa, Israel \\
  \texttt{bron@cs.technion.ac.il} \\
}

\begin{document}
\newcommand{\todo}[1]{{\color{green}[TODO: #1]}}
\newcommand{\daniel}[1]{{\color{blue}[DANIEL COMMENT: #1]}}
\newcommand{\eyal}[1]{{\color{red}[EYAL COMMENT: #1]}}
\newcommand{\alex}[1]{{\color{cyan}[ALEX COMMENT: #1]}}
\newcommand{\parashort}[1]{\noindent \textbf{#1}}

\maketitle

\begin{abstract}
Localization of an object within an image is a common task in medical imaging.  Learning to localize or detect objects typically requires the collection of data that has been labelled with bounding boxes or similar annotations, which can be very time consuming and expensive.  A technique that could perform such learning with much less annotation would, therefore, be quite valuable.  We present such a technique for localization with limited annotation, in which the number of images with bounding boxes can be a small fraction of the total dataset (e.g. less than 1\%); all other images only possess a whole image label and no bounding box.  We propose a novel loss function for tackling this problem; the loss is a continuous relaxation of a well-defined discrete formulation of weakly supervised learning and is numerically well-posed.  Furthermore, we propose a new architecture which accounts for both patch dependence and shift-invariance, through the inclusion of CRF layers and anti-aliasing filters, respectively.  We apply our technique to the localization of thoracic diseases in chest X-ray images and demonstrate state-of-the-art localization performance on the ChestX-ray14 dataset \cite{8099852}.
\end{abstract}

\section{Introduction}
\label{sec:introduction}
Large-scale labelled datasets are one of the key ingredients in the rapidly developing domain of computer-aided diagnosis systems.  In particular, deep learning, which has come to dominate the field of medical imaging in the same way that it has taken over computer vision more generally, is a very data-hungry approach.  The combination of large labelled datasets with deep learning techniques has resulted in state-of-the-art (SOTA) algorithms in many medical imaging tasks, including in classification, detection, and segmentation.

A problematic aspect of this approach is the cost of labelling, particularly in localization tasks -- either detection or segmentation.  In detection, one wishes to find an object in an image by placing a bounding box around it; in the more fine-grained segmentation, one wishes to localize an object with pixel-level granularity.  Generally, in order to use deep learning to train networks to perform either of these tasks in standard fashion, one requires a fair amount of annotated images that mirrors the desired output: bounding boxes in the case of detection, and pixel-level masks in the case of segmentation. The major problem is that collecting such annotations can be very expensive.  Indeed, these annotations are much more expensive than their counterparts in the corresponding classification task, in which the labeller must simply specify a label for the image.  This problem plagues the general computer vision problems of detection and segmentation but is even worse in the case of medical imaging: the labeller must generally be a physician, resulting in a very costly labelling procedure.

Our goal in this paper is to learn to perform localization with considerably less annotation. In particular, we consider the following setting: relatively cheap whole image (i.e. classification-style) labels are available for each image in the dataset, but only a very small number of examples have a bounding box or segmentation mask labels.  This sort of weakly supervised task has been studied before in the computer vision literature \cite{babenko2008multiple}.  In medical imaging, Li \emph{et al.} \cite{Li_2018_CVPR} proposed an approach in the spirit of multiple instance learning and achieved SOTA results on the ChestX-ray14 dataset \cite{8099852}.

Despite the success of the approach in \cite{Li_2018_CVPR}, it has a number of shortcomings, both in terms of the underlying probabilistic model -- which, for example, assumes patch independence, as well in terms of the numerical problems that arise from the formulation.  We propose a new technique in which these problems are resolved.  In particular, our contributions are twofold: First, we propose a novel loss function, which is a continuous relaxation of a well-defined discrete formulation of weakly supervised learning, and which is numerically well-posed; and secondly, we propose a new architecture which accounts for both patch dependence and shift-invariance, through the inclusion of CRF layers and anti-aliasing filters, respectively. Using the new technique, we show SOTA results for the localization task on the ChestX-ray14 dataset.

The remainder of the paper is organized as follows.  Section \ref{sec:related_work} reviews related work.  Section \ref{sec:localization} describes the new technique, focusing alternately on the new loss function and the proposed architecture.  Section \ref{sec:results} presents the experimental results, and Section \ref{sec:conclusions} concludes the paper.

\begin{figure}[h]
 \centering
 \begin{subfigure}[b]{0.32\textwidth}
  \includegraphics[width=1\textwidth]{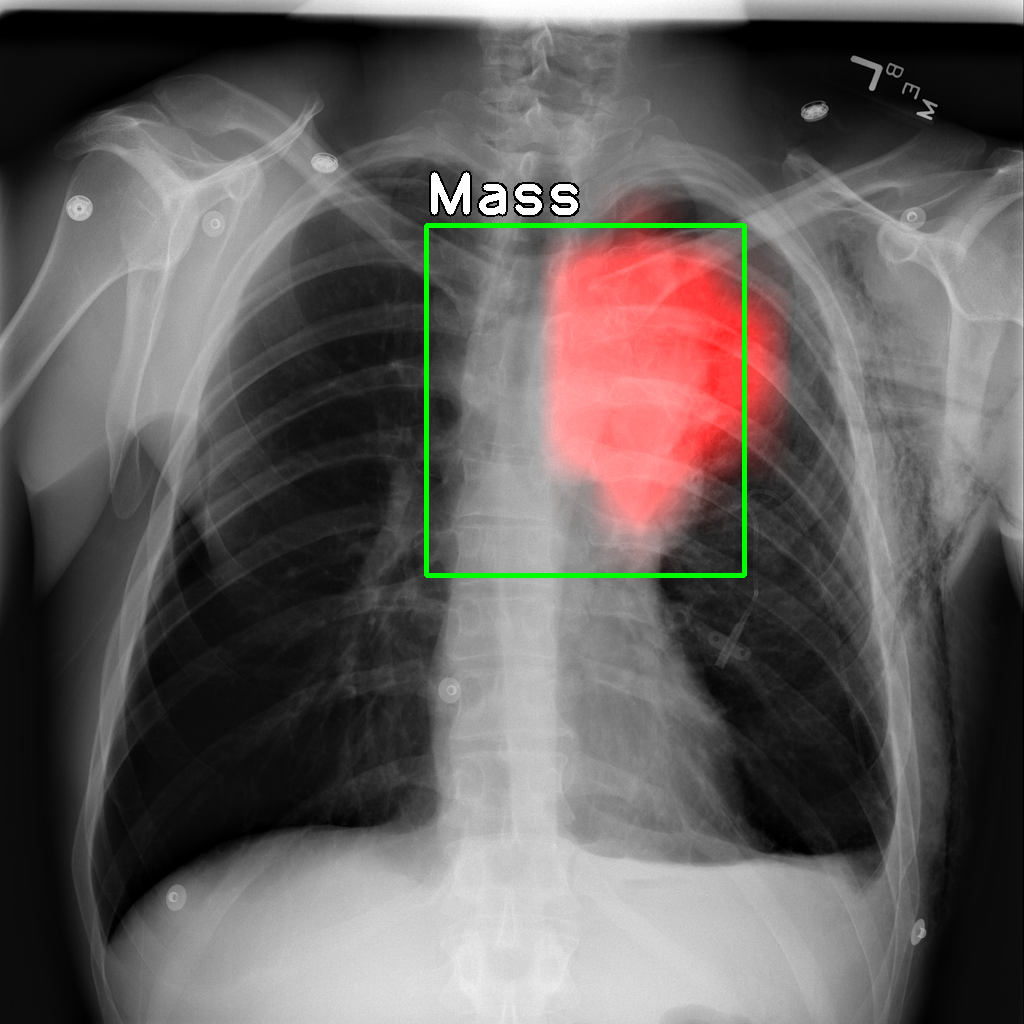}
 \end{subfigure}
 \begin{subfigure}[b]{0.32\textwidth}
  \includegraphics[width=1\textwidth]{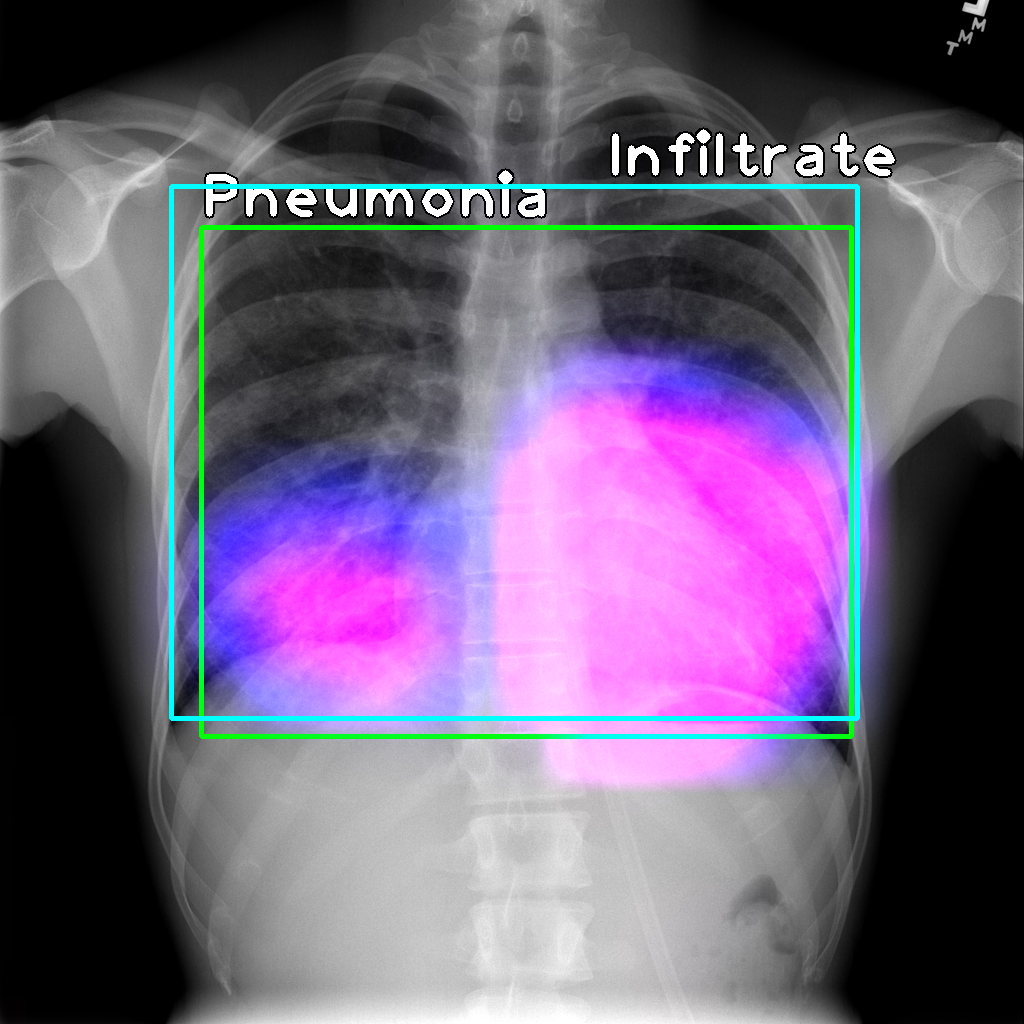}
 \end{subfigure}
 \begin{subfigure}[b]{0.32\textwidth}
  \includegraphics[width=1\textwidth]{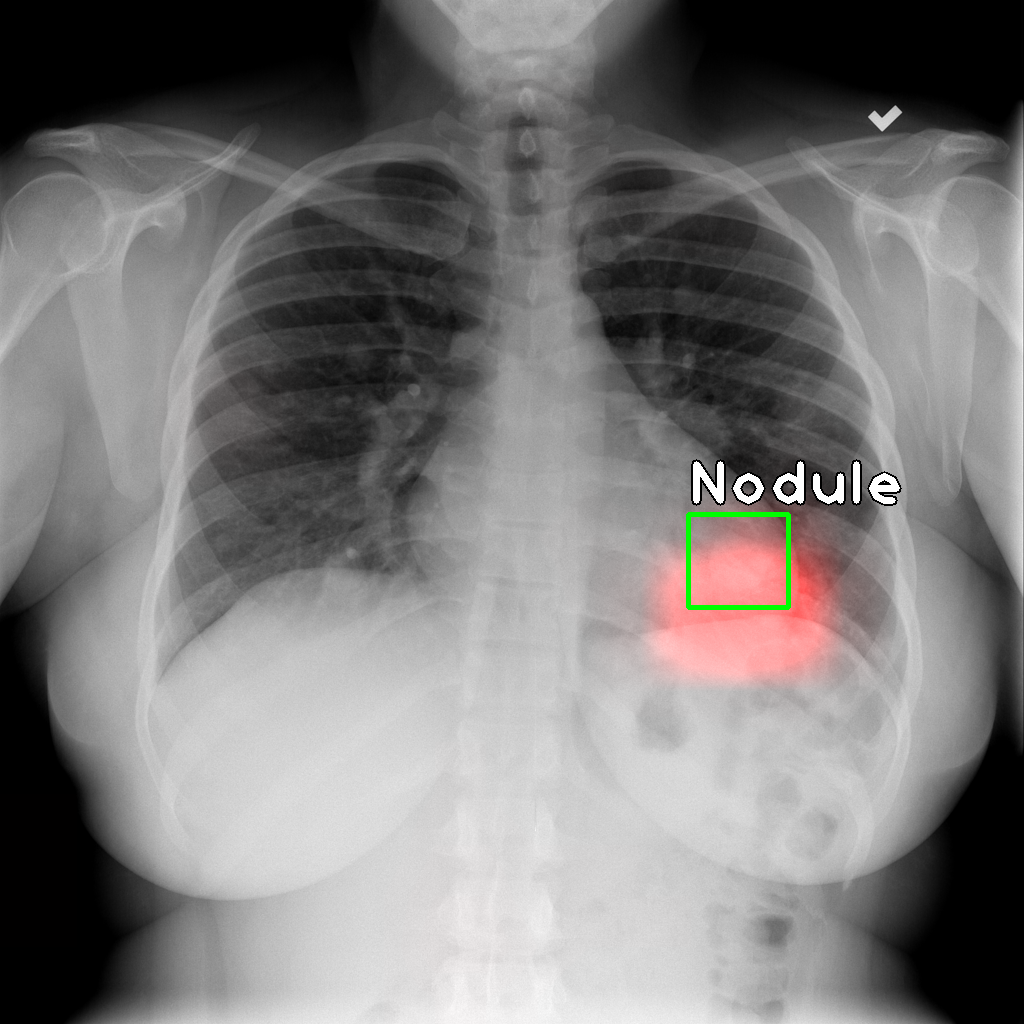}
 \end{subfigure}
 \caption{\small Examples of localized pathologies on test images. The colored blob is the result of our localization algorithms, while the ground truth is marked by a bounding box. In the middle image, two diseases are present, and therefore it is marked twice.}
 \label{fig:qualitative}
 \vspace{-0.5cm}
\end{figure}

\section{Related Work}
\label{sec:related_work}
As our results are presented on the ChestX-ray14 dataset of Wang \emph{et al.} \cite{8099852}, we begin with a brief discussion thereof.  The dataset is a collection of over 100K front-view X-ray images.  Using automatic extraction methods from the associated radiological reports by natural language processing, each image was labelled with up to 14 different thoracic pathology classes; in addition, a small number of images with pathology were annotated with hand-labeled bounding boxes for a subset of 8 of the 14 diseases.

Simultaneous with the release of the dataset, Wang \emph{et al.} \cite{8099852} also presented the first benchmark for classification and localization by weakly-supervised CNN architectures pre-trained on ImageNet.  This benchmark used only the whole label images for training, ignoring the bounding box annotations.  Following the release of the dataset and initial benchmark, several notable works proposed more sophisticated networks for more accurate classification or localization results. Yao \emph{et al.} \cite{Yao2018LearningTD} leveraged the inter-dependencies among all 14 diseases using long short-term memory (LSTM) networks without pre-training, outperforming Wang \emph{et al.} \cite{8099852} on 13 of 14 classes. \cite{Rajpurkar2017CheXNetRP} proposed classifying multiple thoracic pathologies by transfer-learning with fine-tuning, using a 121-layer Dense Convolutional Network (DenseNet) \cite{Huang2016DenselyCC}, yielding SOTA results for the classification task for all 14 diseases. Unlike both previous methods, which do not exploit any of the bounding box annotations, Li \emph{et al.} \cite{Li_2018_CVPR} took advantage of these annotations to simultaneously perform disease identification and localization through the same underlying model. Although their method did not attain the top classification results, they did achieve a new SOTA for localization. Subsequent works \cite{Guan2018DiagnoseLA, 10.1007/978-3-030-13469-3_88} have focused on improving the SOTA in classification; by contrast, our focus is localization rather than classification, so we shall not elaborate further on these results.

Another generally related area of interest is object detection.  Current SOTA object detectors are of two types: two-stage or single-stage.  The two-stage family is represented by the R-CNN framework \cite{Girshick_2014_CVPR}, comprised of a region proposal stage followed by the application of a classifier to each of these candidate proposals. This architecture, through a sequence of advances \cite{girshick2015fast,ren2015faster,he2017mask}, consistently achieves SOTA results on the challenging COCO benchmark \cite{10.1007/978-3-319-10602-1_48}. The initial single-stage detectors, such as YOLO \cite{Redmon_2016_CVPR} and SSD \cite{10.1007/978-3-319-46448-0_2}, exhibited greater run-time speed at the expense of some accuracy.  More recently Lin \emph{et al.} proposed RetinaNet \cite{Lin2017FocalLF}, whose training is based on the ``focal loss''; this network was able to match the speed of previous single-stage detectors while surpassing the accuracy of all existing SOTA two-stage detectors.  These detection approaches, however, are not aimed for tasks that contain a small number of annotated samples as in our setting of interest and are often prone to low accuracy on such datasets.

Finally, we mention the multiple instance learning (MIL) literature \cite{babenko2008multiple}.  Examples of MIL in medical imaging include \cite{yan2016multi,zhu2017deep,hou2016patch}.

\section{Localization with Limited Annotation}
\label{sec:localization}
\subsection{Problem Formulation}
\label{sec:formulation}

\parashort{Setup: The Base Model}
As our starting point, we take the approach of Li \emph{et al.} \cite{Li_2018_CVPR}, which proposes a technique for the classification and localization of abnormalities in radiological images. This approach is very appealing in that it allows for localization to be achieved with a very limited number of bounding box annotations. We now give a brief summary of the technique.

\begin{figure}
  \centering
  \fbox{\includegraphics[width=10cm]{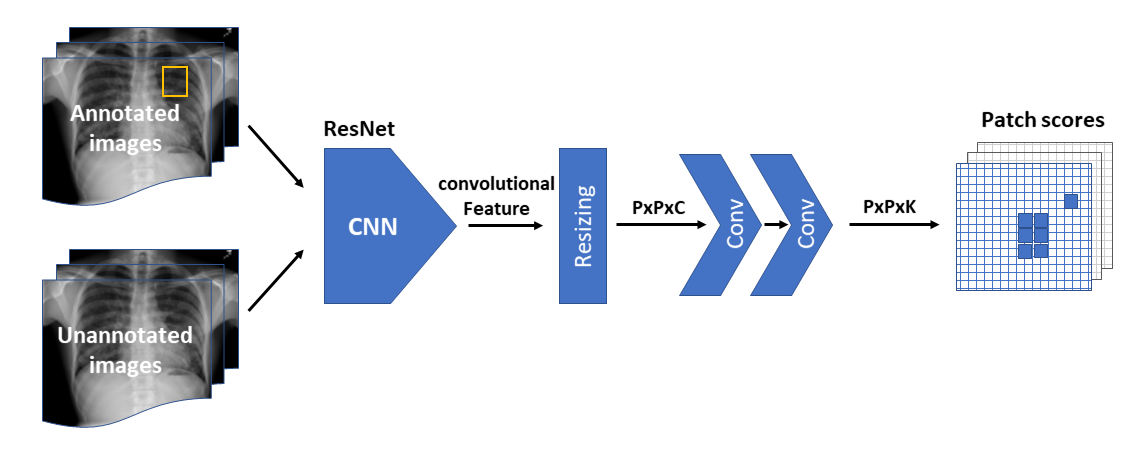}}
  \caption{\small Base model overview \cite{Li_2018_CVPR}. Input images are processed a CNN, extracting their feature maps. The latter are then resized and processed by two subsequent convolutional layers to finally output a \(P \times P\ \times K\) tensor of patch scores.}
  \label{fig:Li's model}
  \vspace{-0.5cm}
\end{figure}

The architecture used in \cite{Li_2018_CVPR} is shown in Figure \ref{fig:Li's model}.  A preact-ResNet network \cite{10.1007/978-3-319-46493-0_38}, with the final classification layer and global pooling layer removed, is used as the backbone; this part of the architecture encodes the images into a set of feature maps. These feature maps are subsequently divided into a \(P\times P\) grid of patches.  Through an application of two convolutional layers (including batch normalization and ReLU activation), the number of channels is modified to $K$,  where \(K\) is the number of possible disease types.  A per-patch probability for each disease class is then derived by the application of a sigmoid function; this is denoted \(p_j^k\), where the probability is that the $j^{th}$ patch of the image belongs to class $k$.  Note that a sigmoid function is applied, rather than a softmax, as a particular patch may belong to more than one disease.

As mentioned above, it is assumed that some images have bounding box annotations, while most do not.  Let us define some terms: the image is $x$; for a disease $k$, the label $y^k = 1$ if the disease is present, $y^k = 0$ otherwise; if the disease $k$ is annotated with a bounding box $b^k$, then $a^k = 1$, otherwise $a^k = 0$.  Now, the loss function can then be broken into two cases, in terms of whether a bounding box annotation is supplied or not.  In the case in which there is a bounding box for a disease of class $k$, i.e. $a^k = 1$, the bounding box $b^k$ is a subset of the $P^2$ patches.  Then the annotated loss is taken to be
\[
L_{ann}^k = -\log p(y^k=1 | x, b^k)
\]
where $p(y^k=1 | x, b^k)$ denotes the probability that disease $k$ is within bounding box $b^k$ of image $x$, and is given by
\begin{equation}
    p(y^k=1 | x, b^k) = \prod_{j \in b^k} p_j^k \prod_{j \in \bar{b}^k} (1-p_j^k)
    \label{eq:ann}
\end{equation}
and $\bar{b}^k$ is the complement of bounding box $b^k$.  The above formula is simply the standard formula for combining independent patch probabilities.  In the case in which no bounding box is supplied, i.e. $a^k=0$, the unannotated loss is
\[
L_{un}^k = -y^k \log p(y^k=1 | x) -(1-y^k) \log (1-p(y^k=1 | x))
\]
where
\begin{equation}
    p(y^k=1 | x) = 1 - \prod_j (1-p_j^k)
    \label{eq:un}
\end{equation}
The latter probability is simply the probability that there is at least one patch with disease $k$, again assuming independence of patches.  Finally, the overall loss per image is
\begin{equation}
    L = \sum_k \left( \lambda_{ann} \,a^k L_{ann}^k \,\, + \,\, (1-a^k) L_{un}^k \right)
    \label{eq:full_loss_old}
\end{equation}

We refer to this model -- the architecture and the loss -- as the \emph{base model}.  It was shown to attain SOTA performance in terms of localization on the NIH Chest X-ray dataset \cite{8099852}.

\parashort{Issues with the Approach of Li \emph{et al.}}
The probabilistic formulation of Li \emph{et al.} is a nice approach to localization tasks with a very limited number of bounding box annotations.  In spite of this, there are several issues with the technique that limit its performance:
\begin{enumerate}
    \item \textbf{Single Patch $\Rightarrow$ Positive Declaration:} In Equation (\ref{eq:un}) of the above derivation for the unannotated loss, only a single patch needs to be positive for a positive disease detection within the image.  In general, this assumption is prone to false positives. One would like for multiple patches to be present for a declaration; in particular, a single positive detection could easily be caused by noise. We would, therefore, like to do away with this assumption, by using a novel loss function.
    \item \textbf{Numerical Issues:} The paper refers to a particular numerical problem that results from the multiplication of many small numbers, as in Equation (\ref{eq:ann}).  This numerical underflow is fatal to the approach, and the problem is circumvented in \cite{Li_2018_CVPR} through a series of unjustified heuristics.  We propose a formulation of the loss in which these issues never arise.
    \item \textbf{Patch Independence:} In both Equations (\ref{eq:ann}) and (\ref{eq:un}), the probabilities of each patch containing an object of a particular class (\(p_j^k\)) are treated as independent between patches. This is not correct in practice as we would like to integrate more elaborate terms that model contextual relationships between object/patch classes. We solve this through the use of a Conditional Random Field (CRF) model.
    \item \textbf{Lack of Shift-Invariance:} As has been pointed out by Zhang \cite{Zhang2019MakingCN}, modern CNNs are not technically shift-invariant.  In order to improve the performance of the localization, one can, therefore, address this issue through the addition of anti-aliasing filters prior to downsampling, as suggested in \cite{Zhang2019MakingCN}.
\end{enumerate}

\parashort{Summary of the Proposed Approach.}
To summarize, our technique is related to the base model but is differentiated in two key ways:
\begin{enumerate}
    \item The use of a novel loss function, which addresses many of the aforementioned issues.
    \item Modifications to the architecture, specifically (a) the incorporation of conditional random field (CRF) layers and (b) the inclusion of anti-aliasing filters.
\end{enumerate}
We now elaborate on each of these, in turn.

\subsection{The New Loss Function}

\parashort{Notation}
As described above, the output of the base model is a tensor of shape the of \(P\times P\times K\); we will continue to denote the output as $p_j^k$, the probability is that the $j$-th patch of the image belongs to class $k$. This output will then be fed into a series of layers that implement a CRF model, with $p_j^k$ representing the unary terms in the CRF.  The output of the CRF is denoted as $z_j^k$.  We will discuss the details of the CRF model in Section \ref{sec:architecture}; for now, we may think of $z_j^k$ as a sharper estimate of whether a particular disease $k$ is present in patch $j$.  $z^k$ indicates the length $P^2$ vector of all patch values for a given disease $k$.

\parashort{The Loss Function: First Pass}

We first consider an annotated example with $a^k=1$, i.e. one with a bounding box. As above, the bounding box $b^k$, consisting of a subset of the patches, contains an object of class $k$.  In this case, the following loss function is natural:
\begin{align}
    L_{ann}^k = - \mathbb{I}[&z^k \text{ contains a blob of size} \ge \tau^k|b^k| \text{ within } b^k \notag \\
    &\text{ and contains blobs of total size} \le \rho^k|\bar{b}^k| \text{ within } \bar{b}^k]
\end{align}
where $\mathbb{I}[\cdot]$ is the indicator function; $\tau^k, \rho^k \in [0,1]$ are thresholds; and $\bar{b}^k$ is the complement of the bounding box $b^k$.  A blob may be made precise as a connected component; however, this will not lead to a nice differentiable loss. So we make the following continuous relaxation of the above discrete formulation:
\begin{equation}
    L_{ann}^k = - \sigma(\mathbf{1}^T (\mathbb{I}[b^k] \odot z^k) - \tau^k|b^k|) \,\cdot\, \sigma(\rho^k|\bar{b}^k| - \mathbf{1}^T (\mathbb{I}[\bar{b}^k] \odot z^k))
    \label{eq:loss1}
\end{equation}
where $\mathbb{I}[b^k]$ is now an indicator function on the bounding box; $\odot$ is the Hadamard product; $\mathbf{1}$ is the vector of all $1$'s; and $\sigma$ is a sigmoid function, i.e. a smooth approximation to the indicator function.  What this says is that there must be a total of $\tau^k|b^k|$ patches within $b^k$ which detect class $k$; the relaxation is that the total no longer has to be in a single connected component.  This is a reasonable relaxation, especially since the CRF already encourages smoothness. In addition, we require that there be fewer than $\rho^k|\bar{b}^k|$ patches outside of $b^k$ which detect class $k$.

Note that this logic extends in a straightforward manner to the case of an unannotated example with $a^k=0$, when there is no bounding box specified. In the case of a positive patch (i.e. one in which disease $k$ is present), the loss is simply
\begin{equation}
    L_{un|pos}^k = - \sigma(\mathbf{1}^T z^k - \hat{\tau}^k)
    \label{eq:loss2}
\end{equation}
so that the threshold $\hat{\tau}^k$ now has a meaning in absolute terms, i.e. the absolute number of patches vs. the number of patches relative to the size of a bounding box. In the case of a negative patch -- where disease $k$ is absent -- we have an equation analogous to (\ref{eq:loss2}):
\begin{equation}
    L_{un|neg}^k = - \sigma(\hat{\rho}^k - \mathbf{1}^T z^k)
    \label{eq:loss3}
\end{equation}
where $\hat{\rho}^k$ is another threshold, whose meaning is in terms of the absolute number of patches, similar to $\hat{\tau}^k$.  Finally, we can combine Equations (\ref{eq:loss2}) and (\ref{eq:loss3}) to get
\begin{equation}
    L_{un}^k = -y^k \sigma(\mathbf{1}^T z^k - \hat{\tau}^k) -(1-y^k) \sigma(\hat{\rho}^k - \mathbf{1}^T z^k)
    \label{eq:loss4}
\end{equation}

\parashort{Addressing Issues (1) and (2).} The above formulation addresses Issues (1) and (2) raised in Section \ref{sec:formulation}.  Regarding Issue (1), Equation (\ref{eq:loss2}) requires more than a single patch in order to make a positive declaration; the number of patches must be equal to $\hat{\tau}^k$, which is a per-class parameter which can be chosen. Regarding Issue (2), neither Equations (\ref{eq:loss1}) or (\ref{eq:loss4}) involve the multiplication of many small values; indeed, both are well-posed from a numerical point of view.

\parashort{Dealing with Vanishing Gradients} Due to the presence of sigmoid functions in Equations (\ref{eq:loss1}) and (\ref{eq:loss4}), in practice, we experience issues of vanishing gradients during training.  We propose the following remedy, based on a different relaxation.   We replace Equation (\ref{eq:loss1}) with
% \begin{equation}
%     L_{ann}^k = \mathbf{\frac{|\bar{b}^k|}{|b^k|}}\,\,\text{ReLU}(\tau^k|b^k| - \mathbf{1}^T (\mathbb{I}[b^k] \odot z^k)) \,+ \, 
%     \text{ReLU}(\mathbf{1}^T (\mathbb{I}[\bar{b}^k] \odot z^k) - \rho^k|\bar{b}^k|)
%     \label{eq:loss1.2}
% \end{equation}
\begin{equation}
    L_{ann}^k = |b^k|^{-1}\,\,\text{ReLU}(\tau^k|b^k| - \mathbf{1}^T (\mathbb{I}[b^k] \odot z^k)) \,+ \, 
    |\bar{b}^k|^{-1}\,\,\text{ReLU}(\mathbf{1}^T (\mathbb{I}[\bar{b}^k] \odot z^k) - \rho^k|\bar{b}^k|)
    \label{eq:loss1.2}
\end{equation}
Note that there are three fundamental differences between Equations (\ref{eq:loss1}) and (\ref{eq:loss1.2}).  First, we have replaced the sigmoid functions $\sigma$ with ReLU functions; this has the effect of still leading to minimal loss (in this case, zero) once the constraints are satisfied, but leads to more nicely behaved gradients.  Second, we have replaced the multiplication with addition.  Once sigmoids have been replaced by ReLU's, the notion of a ``fuzzy and'' relaxation is no longer relevant; in this case, addition makes more sense, and again leads to better-behaved gradients.  Finally, sigmoids are scaled between $0$ and $1$, whereas ReLU's can grow without bound; this necessitates the insertion of a scaling factor of $|b^k|^{-1}$ and $|\bar{b}^k|^{-1}$, to ensure that the two terms in the sum are properly balanced.

%As the \emph{Relu} function has linear dependence upon patch-number ,towards convergence, the ratio between the number of annotated patches relatively to the number of unannotated patches has to be somewhat compensated to maintain balance between the part that deals with positive patches and the part that deals with the negative ones. Thus, to achieve balance the part of the loss function that deals with the positive patches is multiplied with the ratio between patch number, \(\frac{|\bar{b}^k|}{|b^k|}\).

Similarly, for unannotated samples we replace Equation (\ref{eq:loss4}) with:
\begin{equation}
    L_{un}^k = y^k \,\text{ReLU}(\hat{\tau}^k - \mathbf{1}^T z^k) + (1-y^k) \, \text{ReLU}(\mathbf{1}^T z^k - \hat{\rho}^k)
    \label{eq:loss4.2}
\end{equation}
% \begin{equation}
%     L_{un|pos}^k = ReLu(\hat{\tau}^k - \mathbf{1}^T z^k)
%     \label{eq:loss2.2}
% \end{equation}
% \begin{equation}
%     L_{un|neg}^k = ReLu(\mathbf{1}^T z^k - \hat{\rho}^k)
%     \label{eq:loss3.2}
% \end{equation} 

The thresholds $\tau^k,\hat{\tau}^k, \rho^k, \hat{\rho}^k$ can be treated as parameters of the network, that can be optimized during training, or can be considered hyperparameters. Note that the losses described in Equations (\ref{eq:loss1.2}) and (\ref{eq:loss4.2}) do not suffer from any numerical issues.  This is due to the fact that they are not the product of many individual probabilities; rather, they aggregate information across patches in such a way that the resulting loss is numerically stable.

\parashort{Balancing Factors and the Final Loss Function}
There are two sources of data imbalance to account for.  The first is the large imbalance of negative (non-diseased) vs. positive (diseased) examples in the data.  To deal with this, we modify Equation (\ref{eq:loss4.2}) slightly, to read
\begin{equation}
    L_{un}^k = y^k \,\text{ReLU}(\hat{\tau}^k - \mathbf{1}^T z^k) + \gamma (1-y^k)\, \text{ReLU}(\mathbf{1}^T z^k - \hat{\rho}^k)
    \label{eq:loss4.3}
\end{equation}
where $\gamma$ is the ratio of positive to negative examples in the data.  The latter de-emphasizes negative examples relative to the positive examples and follows the practice in \cite{8099852}.

The second form of imbalance we must account for is that between the annotated and unannotated examples; in practice, there are many more unannotated examples.  However, this is already accounted for by the factor $\lambda_{ann}$ in Equation (\ref{eq:full_loss_old}), which is set to a large value.  Combining our own annotated and unannotated losses in Equations (\ref{eq:loss1.2}) and (\ref{eq:loss4.3}), respectively, using Equation (\ref{eq:full_loss_old}), we arrive at the final form of the per-example loss:
\begin{align}
     L(x, y, a, b) = \sum_k \bigg\{ & \lambda_{ann} \,a^k \bigg[ |b^k|^{-1}\,\,\text{ReLU}(\tau^k|b^k| - \mathbf{1}^T (\mathbb{I}[b^k] \odot z^k(x))) \notag \\
     & \quad \quad \quad \,+  |\bar{b}^k|^{-1}\,\,\text{ReLU}(\mathbf{1}^T (\mathbb{I}[\bar{b}^k] \odot z^k(x)) - \rho^k|\bar{b}^k|) \bigg] 
     \label{eq:full_loss} \\
     &  + (1-a^k) \bigg[ y^k \,\text{ReLU}(\hat{\tau}^k - \mathbf{1}^T z^k(x)) + \gamma (1-y^k)\, \text{ReLU}(\mathbf{1}^T z^k(x) - \hat{\rho}^k) \bigg] \bigg\} \notag
\end{align}
where the dependence of the $z$ variables on the image $x$ has been made explicit.

\subsection{Architectural Modifications}
\label{sec:architecture}

\parashort{CRF Model}
As mentioned in Issue (3) in Section \ref{sec:formulation}, we would like to do away with the assumption of patch independence; and indeed, in our derivation of Equation (\ref{eq:full_loss}), we did not make use of such assumptions.  However, to further bolster the dependence between neighboring patches, we introduce a CRF model into our network.  The CRF introduces, in an explicit manner, a spatial dependency between patches.  The effect of the CRF is to increase the confidence for a given patch's predicted label, and thereby to improve localization.  There are several choices amongst neural network-compatible CRFs; we choose the recent pixel-adaptive convolution (PAC) approach of Su \emph{et al.} \cite{su_2019_CVPR}, due to its simplicity and excellent performance. We thus integrate the PAC-CRF modification to our base network and train the model end-to-end.

Given the patch probability outputs $p_j^k$ of the base model, the unary potentials of the CRF are simply taken as the $\psi_j^{(u)}(k) = p_j^k$.  Thus, in the absence of neighbor dependence, the CRF will simply choose $z_j^k = p_j^k$.  Neighbour dependence is introduced through pairwise potentials.  As in \cite{su_2019_CVPR}, we take this potential to be $\psi_{jl}^{(p)}(k_j,k_l)=G(f_j,f_l) W_{k_j k_l}(\xi_j-\xi_l)$, where $f$ are a set of learnable features on the $P \times P$ grid; $G$ is a fixed Gaussian kernel; $\xi_j$ is the pixel coordinates of patch $j$; and $W$ is the inter-class compatibility function, which varies across different spatial offsets, and is also learned.  The pairwise connections are defined over a fixed window \(\Omega\) around each patch.

% For class label \(k\in{1,...,K}\) and patch indexes \(j,l\) at image \(i\), we define the unary potentials as the output of our main model, the \(P\times P\times K\) tensor, feeded to logistic function (Sigmoid function) to maintain a probability patches: \(\psi_{u,ij}(k)=p_{ij}^k\in\Re^K\). The pairwise potentials are defined as: \(\psi_{p,i}(k_j,k_l)={\varkappa}(\mathbf{f}_j,\mathbf{f}_l)\mathbf{W}_{k_j k_l}[\mathbf{p}_j-\mathbf{p}_l]\) while their connections are defined over a fixed window \(\Omega\) around each patch. \(\Omega(\cdot)\) specifies the pairwise connection pattern originated from each patch, \(\varkappa\) is a fixed Gaussian kernel, and \(\mathbf{W}\) models the compatibility-function and allows it to vary across different spatial offsets and it is learned on training. \(\bf{f}\) are learnable features for pairwise potentials and \(\mathbf{p}\) are patch coordinates.

Our training regime for the CRF is as follows.  First, we train the unary terms using the base model until converges. Then, we freeze the base model and train only the PAC-CRF part, using a \(19\times 19\) PAC filter. As our unary model outputs a $P \times P \times K$ tensor, we add to our PAC-CRF model two 2D convolution layers, each followed by a rectified linear unit and batch-normalization, to output a tensor with the same size. 

\parashort{Anti-Aliasing} An important property of any model whose goal is to perform localization or segmentation is that the output of the model should be shift-invariant with regard to its input. However, as Zhang \cite{Zhang2019MakingCN} has noted, standard CNNs use downsampling layers while ignoring sampling theorem, and are therefore not shift-invariant; this is in spite of the fact that CNNs are commonly used as the backbone of many localization/segmentation tasks. To circumvent this problem, an anti-aliasing filter is required prior to every downsampling part of the network. In particular, Zhang \cite{Zhang2019MakingCN} proposed the insertion of a blur kernel as a low-pass filter prior to each downsampling step in the network; and thereby demonstrated an increased accuracy across several commonly used architectures and tasks. Following \cite{Zhang2019MakingCN}, we thus modify the backbone of our base model and integrate such low-pass filters as part of the preact-ResNet network \cite{10.1007/978-3-319-46493-0_38}.  This effectively addresses Issue (4) in Section \ref{sec:formulation}.

\section{Results}
\label{sec:results}
\begin{figure}
  \centering
  \fbox{\includegraphics[width=10cm]{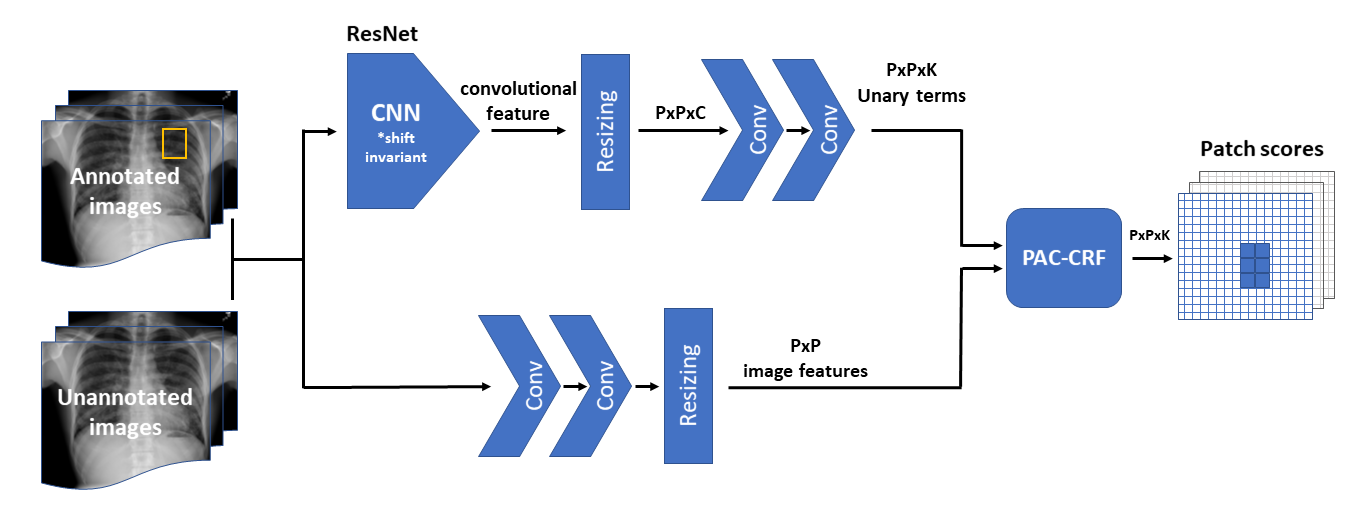}}
  \caption{\small Our model consists of two branches: the upper one has the same architecture as in Figure \ref{fig:Li's model}, though modified with anti-aliasing filters, and computes the unary terms of the CRF model. The lower branch extracts features from the input images to form a feature-tensor of the same size as the unary terms, \(P \times P\), which are used in the pairwise terms. Both enter into the PAC-CRF \cite{su_2019_CVPR}, outputting a \(P \times P \times K\) tensor of patch scores.}
  \label{fig:model}
  \vspace{-0.4cm}
\end{figure}

\parashort{Dataset}
As in \cite{Li_2018_CVPR}, We have examined our model over NIH Chest X-ray dataset \cite{8099852}. The NIH Chest X-ray dataset consists of 112,120 frontal-view X-ray images with 14 disease labels (each image can have multi-labels); images can have no label as well. Out of the more than 100K images, the dataset contains only 880 images with bounding box annotations; some images have more than one such box, so there are a total of 984 labelled bounding boxes.  The remaining 111,240 images have no bounding box annotations but do possess class labels.  The images are \(1024\times1024\), but we have resized them \(512\times512\) for faster processing; we have also normalized the image range to \([-1, 1]\).

The 984 bounding boxes annotations are only given for 8 of the 14 disease types.  Despite this fact, we have noticed that we get superior results by continuing to learn on the full complement of 14 disease types, which seems to imply some interesting interdependence amongst disease classes.

%Even though only 8 classes contain annotations, we have noticed that we can improve diagnose by exploiting dependencies among other labels (diseases) that does not contain any bounding-box annotations; model with 14 classes outperform the one with solely 8, annotated, classes. This may somewhat help to overcome with the general scarcity of the published medical data. The presence of labeled but unannotated class, may point to the presence of other. Thus, training a model to recognize the potential for these interdependencies could enable better prediction across all categories while maximizing the data utilization and its statistical efficiency.

\parashort{Evaluation Metrics}
Although we train over both annotated and unannotated examples, we are interested in localization accuracy, so we evaluate solely over annotated samples.  We use Intersection-over-Union (IoU) and Intersection-over-Region (IoR) to measure localization accuracy. A patch is taken to be positive, i.e. the disease $k$ is present in patch $j$, if its value is greater than 0.5: \(z_j^k\geq 0.5\).  The union of all positive patches is the detected region.  The IoU and IoR can then be computed between the ground truth bounding box and the detected region.  A localization is taken to be correct if $\text{IoU} \ge T$ or $\text{IoR} \ge T$ for given threshold $T$; following the practice of \cite{Li_2018_CVPR}, we use $T=0.1$.  Performance statistics are then computed over 5-fold cross-validation of the annotated samples, as is also done in \cite{Li_2018_CVPR}.

\begin{table}
\scriptsize
\begin{center}
\caption{\small IoU and IoR disease localization accuracy, using 80\% of annotated samples and 0\% or 20\% of unannotated samples (selected randomly) for training. Evaluation set composed of the remaining 20\% annotated-samples of each fold. Showing significant improvement over SOTA results.}
\vspace{0.2cm}
\begin{tabular}{ |c|c||c|c|c|c|c|c|c|c| } 
\hline
\multicolumn{10}{|c|}{IoU accuracy} \\
\hline
\emph{un}[\%] & model & Atelectasis & Cardiomegaly & Effusion & Infiltration & Mass & Nodule & Pneumonia & Pneumothorax \\
\hline
\multirow{2}{1em}{0\%} & ours & \textbf{0.818} & \textbf{1} & \textbf{0.882} & \textbf{0.927} & \textbf{0.695} & \textbf{0.404} & \textbf{0.918} & \textbf{0.726} \\
%\cline{2-10}
& ref. & 0.488 & 0.989 & 0.693 & 0.842 & 0.342 & 0.081 & 0.715 & 0.437 \\ 
\hline\hline
\multirow{2}{1em}{20\%} & ours & \textbf{0.779} & \textbf{1} & \textbf{0.843} & \textbf{0.945} & \textbf{0.709} & \textbf{0.444} & \textbf{0.915} & \textbf{0.732} \\
%\cline{2-10}
& ref. & 0.687 & 0.978 & 0.831 & 0.9 & 0.634 & 0.241 & 0.568 & 0.576 \\ 
%\hline\hline
%\multirow{2}{1em}{xx\%} & \multirow{2}{1em}{ref.} & 0.781 & 0.989 & 0.894 & 0.957 & %0.74 & 0.536 & 0.715 & 0.637 \\
%& & 100\% & 0\% & 100\% & 100\% & 100\% & 80\% & 0\% & 100\% \\ 
\hline\hline
\multicolumn{10}{|c|}{IoR accuracy} \\
\hline
\emph{un}[\%] & model & Atelectasis & Cardiomegaly & Effusion & Infiltration & Mass & Nodule & Pneumonia & Pneumothorax \\
\hline
\multirow{2}{1em}{0\%} & ours & \textbf{0.889} & \textbf{1} & \textbf{0.92} & \textbf{0.95} & \textbf{0.773} & \textbf{0.58} & \textbf{0.933} & \textbf{0.767} \\
%\cline{2-10}
& ref. & 0.528 & \textbf{1} & 0.753 & 0.875 & 0.452 & 0.111 & 0.786 & 0.473 \\ 
\hline\hline
\multirow{2}{1em}{20\%} & ours & \textbf{0.844} & \textbf{1} & \textbf{0.896} & \textbf{0.967} & \textbf{0.808} & \textbf{0.52} & \textbf{0.935} & \textbf{0.806} \\
%\cline{2-10}
& ref. & 0.724 & 0.991 & 0.874 & 0.921 & 0.674 & 0.271 & 0.644 & 0.624 \\ 
\hline
% \bottomrule
\end{tabular}
\label{table:IoX}
\end{center}
% \vspace{-0.5cm}
\end{table}

\parashort{Training Details} We use ResNet-50 as the backbone of our model.  We initialize the weights based on ImageNet \cite{deng2009imagenet} pre-training and then allow them to evolve as training proceeds.   We take \(P=20\) for a $20 \times 20$ patch grid, and take \(\lambda_{ann} = 70\). We use the ADAM optimizer \cite{kingma2014adam} with a weight decay regularization coefficient equal to $0.01$ and exponentially-decaying learning-rate, initialized to $0.001$. We take the batch size to be 48.

%\parashort{Loss Thresholds: \texorpdfstring{\(\tau^k,\hat{\tau}^k, \rho^k, \hat{\rho}^k\)}{Lg}} 
%\parashort{Loss Parameters}

It is natural to treat the thresholds \(\tau^k,\hat{\tau}^k, \rho^k, \hat{\rho}^k\) as parameters of the network, which can be optimized during training.  Unfortunately, this quickly leads to the degenerate solution, i.e: $\tau,\hat{\tau}\rightarrow{0}$, $\rho \rightarrow 1$ and $\hat{\rho}\rightarrow P^2$. In order to avoid reaching the degenerate solution, we employ the following procedure.  We begin by freezing the thresholds, and training only the network weights; we then freeze the network weights, and find the optimal thresholds; we continue to alternate this procedure until convergence.

\parashort{Experiments}
We compare our results with those of Li \emph{et al.} which represent the SOTA for the localization task on the NIH Chest X-ray dataset.  As has been mentioned above, we use 5-fold cross-validation.  We examine two separate settings:  (a) the model is trained using only 80\% of the annotated samples, with the 80\% representing the training part of the fold; (b) the model is trained using 80\% of the annotated samples as described in (a), as well as 20\% of the unannotated samples, representing about 20K samples (selected randomly).  In both cases, the results are evaluated on the remaining 20\% of the annotated samples of each fold.

In Table \ref{table:IoX} we present two versions of the localization accuracy, with the top based on IoU and the bottom based on IoR; best results are shown in bold.  Our method outperforms that of Li \emph{et al.} for all 8 disease classes, for IoU and IoR, and for both settings - with no extra unannotated data added, and with 20\% unannotated data.  Examining the IoU data more closely, we see several patterns.  First, we perform considerably better than \cite{Li_2018_CVPR} when no unannotated data is added; for example, the accuracy on Atelectasis and Mass is nearly \emph{double} that of \cite{Li_2018_CVPR}, whereas the performance on Nodule is \emph{five times} better.  Second, with the addition of unannotated data, the gaps narrows -- for example, Nodule is now slightly less than double, and many other disease classes have quite a bit smaller gap -- but the gap is still present for each disease class.  We hypothesize that our improvement is less in most cases simply because the algorithm trained with no unannotated data already has a fairly high performance in most cases; thus, the marginal benefit of adding the unannotated data is smaller.  Third, in examining our own results, we see that the addition of unannotated data often helps, but does not always do so.  In particular, there is an increase in localization accuracy for four of the eight diseases -- Infiltration, Mass, Nodule, Pneumothorax -- while two diseases, Cardiomegaly and Pneumonia, undergo little or no change.  The remaining two, Atelectasis and Effusion, actually suffer a decrease in accuracy due to the addition of the extra unannotated data.  In examining the data, the solution to this puzzle becomes apparent: Atelectasis and Effusion have the largest number of annotated examples of the eight disease classes.  This explains why they have quite high localization accuracies to begin with, when no unannotated data has been added (0.818 and 0.882, respectively); and why the addition of extra unannotated examples does not help. (It is interesting to note that Pneumonia has high accuracy and a decent number of annotated examples; the reason it differs from Atelectasis and Effusion is that it also has a relatively small number of \emph{unannotated} examples, so that adding them does not affect its performance too much.)

We note that the IoR data is fairly similar to the IoU data, and most of the observations above hold in this case as well.  Qualitative examples of the localizations derived are shown in Figure \ref{fig:qualitative}.

\section{Conclusions}
\label{sec:conclusions}
We have presented a new technique for localization with limited annotation.  Our method is based on a novel loss function, which is mathematically and numerically well-posed; and an architecture that explicitly accounts for patch non-independence and shift-invariance.  We present SOTA results for localization on the ChestX-ray14 dataset.  Future work will focus on applying these ideas to the realm of semantic segmentation.

\label{sec:references}
\bibliography{references}
\bibliographystyle{unsrt}

\end{document}